\documentclass[11pt]{article}

\usepackage[final]{acl}

\usepackage{times}
\usepackage{latexsym}
\usepackage[T1]{fontenc}
\usepackage[utf8]{inputenc}
\usepackage{microtype}
\usepackage{inconsolata}
\usepackage{graphicx}
\usepackage{booktabs}
\usepackage{multirow}
\usepackage{amsmath}
\usepackage{amssymb}
\usepackage{xcolor}
\usepackage{url}

\graphicspath{{figures/}}

\newcommand{\ind}{\mathbf{1}}

\newcommand{\pth}[1]{{\small\path{#1}}}
\newcommand{\pyrun}[1]{{\small\texttt{python\ -m\ }\path{#1}}}

\title{Counterexample Generation via Per-Theorem Symbolic Verifiers:\\
When Imitation Hurts and Reinforcement Repairs}

\author{
  Omar Farouk Zouak\textsuperscript{1}\thanks{\;Corresponding author:
    \texttt{omar.zouak@ensia.edu.dz}} \quad
  Houssam Eddine Boukhalfa\textsuperscript{1} \quad
  Soumaya Lakehal\textsuperscript{1} \\[0.15em]
  \bfseries Shiv Katiyar\textsuperscript{2} \quad
  Samia Nefti-Meziani\textsuperscript{2} \\[0.5em]
  {\normalfont\textsuperscript{1}National School of Artificial Intelligence, Algiers, Algeria} \\
  {\normalfont\textsuperscript{2}University of Birmingham, United Kingdom}
}

\begin{document}
\maketitle

\begin{abstract}
Large language models often solve a theorem forward yet fail to disprove a closely related false one: a falsification gap that supervised fine-tuning does not close and can actively worsen. We frame counterexample generation as constrained witness emission against a deterministic per-theorem Python verifier, and release \textsc{SymCE}, a corpus of 4{,}707 false undergraduate-algebra and real-analysis conjectures, each paired with executable verifiers. The verifier also serves as the reward function, making \textsc{SymCE} a training environment. Training Qwen3-4B with SFT followed by GRPO under this oracle reveals an \emph{imitation trap}: counterexample-only SFT collapses true-theorem recognition from 0.27 to 0.00, while RLVR with a sparse outcome-only reward repairs this and exceeds the base, to 0.66. The collapse replicates across four seeds and on Gemma-3-4B. Sparse and dense rewards yield statistically indistinguishable in-domain success yet diverge by 33 points on a held-out calibration probe, a dissociation we trace to the partial-credit term. Our 4B model outperforms every evaluated 7B open-weights math specialist, remains competitive with six frontier commercial APIs, and transfers under unchanged prompting to GSM8K, MATH-500 and MMLU-college-math. A human audit of 177 verifier decisions finds 97.7\,\% accuracy. Code, data, verifier modules and annotations: \url{https://github.com/ce-rlvr/SymCE}.
\end{abstract}

\section{Introduction}
\label{sec:introduction}

Large Language Models (LLMs) excel at forward mathematical derivation, but struggle significantly with inverse reasoning and generating counterexamples to false claims. Recent benchmarks demonstrate that even frontier models fail at falsification tasks they can easily forward-solve \citep{sinha2025refute, li2025countermath}. Furthermore, standard imitation learning (SFT) often exacerbates this by inducing format-biases rather than true logical reasoning \citep{spurious25}.

The natural fix is to train counterexample generation directly, but two obstacles have so far prevented this from becoming a primary objective.
First, large-scale counterexample data is scarce: the only widely cited counterexample benchmark, COUNTERMATH, contains 1{,}216 problems, and its evaluation is mediated by an LLM judge.
Second, supervised fine-tuning (SFT) on counterexamples appears insufficient.
Counterexample-only training data contains, by construction, only false statements, and we show below that imitation-only training transfers this distribution shift into the policy: a model trained purely to emit counterexamples loses the ability to recognise true statements, and confidently fabricates witnesses on theorems that are in fact true.

We attack both obstacles together. We build \textsc{SymCE}, a deterministic, large-scale counterexample environment, by hypothesis-dropping over 6{,}748 undergraduate-level theorems and generating, for each \texttt{(theorem\_id, dropped\_hypothesis)} pair, an independent Python verifier module that returns one boolean per remaining hypothesis and one for the negated conclusion. The pipeline retains 4{,}707 records. Evaluation is fully symbolic: the policy emits a one line Python expression, and the verifier returns a boolean tuple. Because the verifier runs in tens of milliseconds, the same substrate serves as the reward function during reinforcement learning, so \textsc{SymCE} is a training environment and not only a benchmark.

We use this environment to train Qwen3-4B \citep{qwen3report} with SFT followed by GRPO under two reward shapes and to evaluate against five 7B open-weights math specialists, six commercial-API frontier models, and four external mathematics benchmarks. \emph{In doing so, we uncover a pathology}: imitation-only training on counterexamples collapses the policy's ability to recognise true theorems, and only verifier-grounded RL with an outcome-only reward repairs it. Partial-credit dense rewards yield equal in-domain success but do not restore calibration; a dissociation that, to our knowledge, has not been previously isolated.

\paragraph{Contributions.}
We make four main contributions.
\textbf{(C1) The Imitation Trap Pathology.} We isolate a calibration failure mode in counterexample-only supervised fine-tuning: Pass@1 on true theorems collapses from 0.27 to 0.00 under the training prompt style, because the policy learns the output \emph{form} of a counterexample without the negated-conclusion constraint. The collapse reproduces in all four training seeds and on a second model family.
\textbf{(C2) Verifier-Grounded RLVR for Falsification.} We demonstrate that Reinforcement Learning with Verifiable Rewards (RLVR), guided by a deterministic symbolic execution interface, repairs this pathology. Furthermore, we show that the partial-credit reward term that is the natural dense shaping for this task structure preserves logical calibration markedly worse than an outcome-only reward, and we identify the mechanism.
\textbf{(C3) Strong Empirical Performance \& Transfer.} Our Qwen3-4B RLVR model achieves 0.49\,$\pm$\,0.02 Pass@1 on in-domain falsification across four seeds, outperforming all evaluated 7B open-weights math specialists and remaining competitive with frontier APIs, while demonstrating strong zero-shot transfer to standard forward-math tasks (MATH-500, GSM8K).
\textbf{(C4) \textsc{SymCE}: A Deterministic Falsification Environment.} To enable this training, we introduce and release a programmatically verifiable corpus of 4{,}707 false algebraic conjectures with per-theorem executable verifiers, moving beyond the noisy LLM-as-a-judge paradigm.

\section{Related Work}
\label{sec:related}

\paragraph{Counterexample and falsification benchmarks.}
The closest comparable resource for counterexample reasoning is COUNTERMATH \citep{li2025countermath}, a 1{,}216-problem benchmark of university-level counterexample-driven proofs evaluated with an LLM-as-judge protocol; the largest frontier model reported in that work achieves an F1 of 60.1.
REFUTE \citep{sinha2025refute} measures counterexample success on algorithmic problems and reports a sharp asymmetry between forward solution rates (around 48\,\%) and counterexample success (below 9\,\%) for the same models.
The most directly related training resource is \textit{Learning to Disprove} \citep{li2026learning}, which constructs roughly 575K hypothesis-dropping mutations of Lean 4 Mathlib theorems and uses Lean tactic checks as the verification signal.
Our dataset is smaller than \citeauthor{li2026learning}'s but uses a Python-verifier substrate that is more faithful to the algebraic and real-analytic structures evaluated, runs in tens of milliseconds rather than seconds, and is paired with the witness-schema interface of \S\ref{sec:method:schema}.
The Alchemy line of AST-mutation work \citep{wu2024alchemy} focuses on generating valid mutated \emph{true} theorems for automated theorem proving and is complementary rather than competing.

\paragraph{Verifiable-reward reinforcement learning for mathematics.}
RLVR with GRPO has become the dominant post-training paradigm for forward mathematical reasoning, anchored by DeepSeek-R1 \citep{guo2025deepseekr1} and adopted in PRIME \citep{cui2025prime} and Eurus \citep{yuan2024eurus}. While recent studies debate whether RLVR installs new reasoning or merely elicits latent capabilities via format adaptation \citep{spurious25}, our out-of-distribution transfer results (\S\ref{sec:results:ood}) suggest falsification training installs fundamentally transferable inverse-reasoning skills.

\paragraph{Process supervision and symbolic verifiers.}
Process reward models (PRMs), introduced for mathematical reasoning by \citet{lightman2024verify} and extended by Math-Shepherd \citep{wang2023mathshepherd}, and the FOVER pipeline \citep{kamoi2026fover}, which mines step-level labels from Z3 and Isabelle traces, supply dense supervision at the step level for forward proving.
Our per-theorem verifier sits between an outcome reward model and a PRM: it returns one boolean per remaining hypothesis and one for the negated conclusion, which gives the reward function (Eq.~\ref{eq:reward}) a structured, audited signal without the human-annotation cost of a learned PRM.
PRL \citep{yao2026prl} and self-rewarding correction \citep{xiong2025selfrewarding} bear directly on where such an intermediate signal sits; our boolean tuple differs from both in being externally grounded and non-learned, which is what lets \S\ref{sec:results:reason} attribute a behavioural change to one term of the aggregation.
SymPy and Z3 \citep{meurer2017sympy,demoura2008z3} provide the underlying verification primitives.

\paragraph{The hallucination tax.}
A growing literature describes a calibration pathology induced by standard imitation-style training: TreeCut \citep{ouyang2025treecut} demonstrates that LLMs continue to fabricate answers when problems are systematically modified to remove a critical constraint; SUM \citep{song2025hallucination} and related work report refusal-rate collapses of roughly 80\,\% under standard rejection-sampling fine-tuning on unanswerable-math benchmarks.
Our imitation-trap finding (\S\ref{sec:results:trap}) is the cleanest measurement of this pathology in the counterexample-generation setting we are aware of, and the first to localise the mechanism, via the reason taxonomy of \S\ref{sec:results:reason}, to the partial-credit term of the reward function.

\section{Dataset and Verification Environment}
\label{sec:method}

We frame counterexample generation as a constrained generation task: given a false conjecture of the form ``if $H_1 \wedge \cdots \wedge H_k$ then $Q$'', output a witness $w$ such that all remaining hypotheses hold and $Q$ fails on $w$. Our environment is built around three design choices: a per-theorem Python verifier compiled once and reused across all rollouts, a structured witness-schema interface that constrains the policy's output shape, and a conclusion-weighted dense reward whose sparse counterpart we use for ablation.

\subsection{Source theorems and hypothesis dropping}
\label{sec:method:mutation}

We collected 6{,}748 unique undergraduate-level mathematical theorems spanning six subdomains (Group, Ring, Field and Number Theory, Linear Algebra and Real Analysis). Each theorem is stored in a structured form with explicit \textit{hypotheses} and \textit{conclusion} fields. To construct false conjectures we drop one hypothesis at a time, giving 18{,}262 candidates. Removing a hypothesis does not guarantee the result is false or even well-posed --- the hypothesis may be redundant or merely clarificatory --- so we apply an LLM-driven filter (Gemini 3.1 Pro): the teacher must classify the mutated theorem as well-posed and non-trivial, emit a valid Python verifier, and propose a counterexample that evaluates to \texttt{True} on that verifier without execution errors, with up to 5 retries under error-trace feedback. The teacher produces a verifier for 12{,}376 candidates; gating then distils these to 4{,}707 verifier-backed records over 3{,}485 unique theorems, of which 4{,}700 carry a witness schema (\S\ref{sec:method:schema}). We report per-subdomain counts in Table~\ref{tab:dataset}.

A sandboxed subprocess imports the verifier dynamically, builds the witness-evaluation namespace, evaluates the policy's witness expression string in that namespace, and calls \texttt{conditions(witness)}, which returns a list of booleans, one per remaining hypothesis followed by one for the negated conclusion. The policy and the oracle share \emph{no code paths} beyond Python's \texttt{eval}; the verifier is fixed at the start of training and reused across all rollouts for that conjecture.

This is a deliberate departure from prior counterexample-evaluation pipelines: COUNTERMATH \citep{li2025countermath} relies on an LLM-as-judge protocol, and \textit{Learning to Disprove} \citep{li2026learning} relies on a Lean tactic check. A Python module evaluated against a witness expression is intermediate: it is symbolic and decidable for the subdomains we target, fast (\S\ref{sec:method:audit}), and reproducible.

\subsection{Witness-schema interface}
\label{sec:method:schema}

We attach to each prompt a one-line \textit{witness schema}: a short Python-shape description of an acceptable witness. Schemas are generated post-hoc by reading the verifier module and its canonical witness with a second teacher model (DeepSeek) and asking for a single-sentence description of the expected Python object shape, with one or two concrete examples. Sample schemas are shown in Appendix~\ref{sec:appendix:schema}.

Schemas are merged into the user prompt. We view this as a small but load-bearing methodological contribution: a structural prior on the output shape, generated automatically from the verifier and a canonical witness, is what allows for the practical RLVR effectiveness without altering either the reward function or the verifier.

\subsection{Split construction}
\label{sec:method:canonical}

We construct three splits with per-domain stratified sampling and a fixed seed (42): the held-out test set (n=201), the SFT split (n=1{,}000), and the RLVR prompt set (n=3{,}506). Table~\ref{tab:dataset} summarises the splits.

\subsection{Reward function}
\label{sec:method:reward}

The per-rollout reward computed by the oracle is
\begin{equation}
\label{eq:reward}
\begin{aligned}
r =\;& \alpha \cdot \frac{s_{\text{hyp}}}{n_{\text{hyp}}} + \beta \cdot q_{\neg} + \gamma \cdot \ind[\text{format ok}] \\
&{-} \delta \cdot \ind[\text{gave up on false}] \\
&{-} \varepsilon \cdot (1 - \ind[\text{parse ok}])\,,
\end{aligned}
\end{equation}
where $s_{\text{hyp}} = \sum_{i=1}^{n_{\text{hyp}}} \texttt{conditions}[i]$ counts satisfied hypotheses, $q_{\neg} = \texttt{conditions}[\texttt{neg\_q\_index}] \in \{0, 1\}$ is the negated-conclusion indicator, and the format / parse flags come from the structural and witness-evaluation checks. The penalty $\delta$ applies when the policy emits the literal sentinel \texttt{NONE}. Every training conjecture is false, so this is always a give-up signal.

\paragraph{Dense and sparse variants.} We study two reward shapes. \textit{Dense}: $\alpha{=}0.25, \beta{=}0.65, \gamma{=}0.10, \delta{=}0.30, \varepsilon{=}1.00$, range approximately $[-1, +1]$. \textit{Sparse}: $\alpha{=}\gamma{=}\delta{=}0, \beta{=}\varepsilon{=}1$, range $\{-1, 0, +1\}$. The same SFT checkpoint is used as the reference policy for both.


\paragraph{Conclusion-weighting bound.} A structural property of Eq.~(\ref{eq:reward}) under dense weights is that any successful, formatting-compliant rollout ($\ind[\text{format ok}] = \ind[\text{parse ok}] = 1$ and $\ind[\text{gave up on false}] = 0$) that falls into the ``hypothesis-only'' failure mode ($s_{\text{hyp}} = n_{\text{hyp}}$, $q_{\neg} = 0$) is strictly dominated by any valid rollout that successfully negates the conclusion ($q_{\neg} = 1$):
\[
r_{\text{hyp-only}} \;=\; \alpha + \gamma \;<\; \beta + \gamma \;\leq\; r_{\text{neg-Q}}.
\]
This structurally precludes the ``all hypotheses satisfied but conclusion asserted true'' failure mode that affects naïve formulations \citep{li2026learning}.

\subsection{Verifier audit and operational measurements}
\label{sec:method:audit}

The verifiers are produced by a teacher model (Gemini 3.1 Pro) and therefore must be audited before they can support a reward signal. We audit the accept/reject \emph{decision} on real policy outputs rather than the verifier source. The generated verifiers decide conditions with heuristics and finite simplifications that are theoretically incomplete yet correct on the instances that arise, so a judgement on the source is slow to make and largely subjective; the decision is also the quantity the reward signal actually depends on. This audit replaces a hand-curated unit-test suite over the verifier modules, which we abandoned for that reason. We sampled 177 model-generated witnesses from the held-out test split across seeds and training configurations (101 accepted by the verifier and 76 rejected) and had an author with graduate-level training in the relevant areas judge independently whether each is genuinely a valid counterexample. Of the accepted, 99/101 were confirmed valid (false-accept rate 2.0\,\%); of the rejected, 74/76 were confirmed invalid, for 97.7\,\% decision accuracy. \emph{No case was traced to an incorrect verifier module}, and we found no evidence of reward hacking through verifier acceptance. A single study thus speaks to verifier trustworthiness, hypothesis-drop label noise, reward hacking and witness-schema quality at once. Composition and the four disputed decisions are in Appendix~\ref{sec:appendix:audit}; the annotations are released.

Operationally, the sandbox is fast enough to support RLVR at our budget: across the held-out test set with each record's canonical witness fed through the oracle, the median wall-clock per call is 132~ms, p95 145~ms, and the maximum 326~ms. No test record approached the eight-second sandbox timeout.

\begin{table}[t]
\centering
\small
\setlength{\tabcolsep}{4pt}
\begin{tabular}{lrrrr}
\toprule
Domain & Strict & SFT & RLVR & Test \\
\midrule
GroupTheory   & 874 & 366 & 434 & 74 \\
RingTheory    & 565 & 237 & 280 & 48 \\
NumberTheory  & 392 & 164 & 195 & 33 \\
LinearAlgebra & 233 & 98  & 115 & 20 \\
FieldTheory   & 166 & 70  & 82  & 14 \\
RealAnalysis  & 159 & 67  & 79  & 13 \\
\midrule
\textbf{Allocated} & \textbf{2389} & \textbf{1002} & \textbf{1185} & \textbf{202} \\
\bottomrule
\end{tabular}
\caption{Domain-stratified \emph{allocation} of the strict pool: the 2{,}389 records whose teacher-generated canonical witness passes its own verifier at generation time. Rounding the split targets and one oracle-gating drop move three records from SFT and test into the RLVR prompt set, so the \emph{final} split sizes are SFT 1{,}000, test 201, and RLVR $1{,}188 + 2{,}318 = 3{,}506$; the 2{,}318 loose records have a working verifier but no oracle-confirmed canonical witness. Pipeline total: 4{,}707 records over 3{,}485 unique theorems; canonical-witness positive-reward rate 0.98.}
\label{tab:dataset}
\end{table}

\paragraph{Theorem-truth probe.} We construct an auxiliary set of 347 \emph{true} theorems drawn from the same theorem source. The model is asked to classify whether the provided theorem is true or false.

\section{Training Setup}
\label{sec:training}

We train Qwen3-4B \citep{qwen3report} (\texttt{unsloth/Qwen3-4B}) using Supervised Fine-Tuning (SFT) followed by Group Relative Policy Optimization (GRPO) \citep{shao2024deepseekmath}. SFT is performed over 2--3 epochs with a cosine learning rate schedule (Appendix~\ref{sec:appendix:hyperparameters}). For RLVR, we initialise from the SFT adapter and train for 4 epochs ($\approx$\,1{,}752 steps) using a KL penalty of $0.02$ against the SFT reference policy. During RLVR, we sample $G=4$ rollouts per prompt using vLLM \citep{kwon2023vllm} to maintain high-throughput generation. Both training stages use LoRA \citep{hu2022lora} (rank 16, applied to all attention and MLP modules) and bfloat16 precision. 

Notably, our entire pipeline is highly computationally accessible: training executes on a single NVIDIA RTX 5090 (32\,GB) consumer GPU, with SFT completing in under five minutes and each RLVR run taking approximately nine hours. Full hyperparameters, optimizer configurations, and prompt templates are detailed in Appendix~\ref{sec:appendix:hyperparameters}.

\section{Experimental Setup}
\label{sec:experiments}

\subsection{Models}
\label{sec:experiments:models}

We report four model variants from this work: the unmodified \textbf{Qwen3-4B base}, \textbf{SFT} checkpoint, and two RLVR checkpoints \textbf{RLVR-dense} (Eq.~\ref{eq:reward} dense weights) and \textbf{RLVR-sparse} (sparse weights). For external context we evaluate five open-weights 7B math-specialised models, Abel-7B-002, DeepSeek-Math-7B-RL, Eurus-2-7B-PRIME, MAmmoTH2-7B-Plus and Qwen2.5-Math-7B-Instruct \citep{shao2024deepseekmath, yang2024qwen25math, cui2025prime, yue2024mammoth2}, and six frontier commercial APIs from the DeepSeek and Gemini families. Per-model breakdowns of the external comparators are deferred to Appendix~\ref{sec:appendix:external-models}; in the main paper we summarise them as a best-7B row and a frontier-API range. For the cross-family replication of \S\ref{sec:results:robustness} we repeat the pipeline on Gemma-3-4B.

\subsection{Benchmarks}
\label{sec:experiments:benchmarks}

\textbf{Counterexample (in-domain, n=201).} Our held-out test split (\S\ref{sec:method:canonical}), stratified by domain. The strict scoring criterion is $q_{\neg}{=}1 \wedge s_{\text{hyp}}{=}n_{\text{hyp}}$; we also report the relaxed \textit{conclusion-pass} rate $q_{\neg}{=}1$.

\textbf{Theorem-truth (n=347).} The model is considered correct iff it recognises the theorem as true. We evaluate two prompt styles: \textsc{SymCE}-prompt (the same prompt style as training) and \textit{truth-classify} (an explicit true/false question).

A response counts as correct only when the model commits to an answer. Permissive extraction inflates untrained baselines substantially; Appendix~\ref{sec:appendix:scoring} reports both readings and validates the rule.

\textbf{External math benchmarks.} GSM8K (n=1319), MATH-500 (n=500), MMLU-college-math (n=100), AIME-2024 (n=30) \citep{cobbe2021gsm8k, hendrycks2021math, hendrycks2021mmlu}. All four models use \textbf{identical} prompting (standard chain-of-thought); our SFT and RLVR splits contain \textbf{no} examples from these benchmarks and no training on their answer-extraction formats.

\subsection{Decoding and metrics}
\label{sec:experiments:decoding}

We evaluate under two decoding regimes: \textit{greedy} (temperature 0, $n{=}1$) and \textit{sampled} (temperature 0.7, top-$p$ 1.0, $n{=}8$). Pass@$k$ for $k \in \{1,3,5,8\}$ is computed with the unbiased estimator of \citet{chen2021codex}; greedy Pass@1 is reported as the headline number throughout. 95\,\% confidence intervals are percentile bootstrap intervals over problem ids with 10{,}000 resamples (the unbiased Pass@$k$ estimator is applied per problem \textit{before} resampling). Paired deltas (model A minus model B) bootstrap over the per-problem Pass@$k$ differences. All scripts are released.

\section{Results}
\label{sec:results}

\subsection{Counterexample generation}
\label{sec:results:main}

\begin{table*}[t]
\centering
\small
\setlength{\tabcolsep}{3pt}
\begin{tabular}{lccccc}
\toprule
Model & Pass@1 (greedy) & Pass@8 (sampled) & Concl.-pass & Format-OK & Parse-OK \\
\midrule
Qwen3-4B base                        & 0.30 \tiny{[0.24, 0.37]} & 0.49 \tiny{[0.42, 0.56]} & 0.30 & 0.45 & 0.42 \\
\quad + SFT                          & 0.35 \tiny{[0.28, 0.41]} & 0.64 \tiny{[0.57, 0.70]} & 0.41 & \textbf{1.00} & 0.82 \\
\quad + RLVR (dense)                 & \textbf{0.50} \tiny{[0.43, 0.57]} & \textbf{0.68} \tiny{[0.61, 0.74]} & \textbf{0.61} & 0.98 & 0.94 \\
\quad + RLVR (sparse)                & 0.50 \tiny{[0.43, 0.57]} & 0.67 \tiny{[0.61, 0.74]} & 0.59 & 0.97 & \textbf{0.95} \\
\midrule
Best open-weights 7B (Eurus-2-7B-PRIME) & 0.30 \tiny{[0.24, 0.36]} & 0.57 \tiny{[0.50, 0.64]} & 0.34 & 0.87 & 0.70 \\
Frontier APIs (range over 6)\textsuperscript{a} & 0.42\,--\,0.58 & up to 0.77 & --- & --- & --- \\
\bottomrule
\end{tabular}
\caption{In-domain counterexample benchmark (n=201), reference seed 42. Greedy Pass@1 is the headline metric; 95\,\% bootstrap CIs in brackets. \textit{Conclusion-pass} is the relaxed criterion $q_{\neg}{=}1$ of \S\ref{sec:experiments:benchmarks}. Four-seed means for the trained variants are in Table~\ref{tab:multiseed}. \textsuperscript{a}Range across six DeepSeek/Gemini frontier APIs. The full per-model open-weights table is in Appendix~\ref{sec:appendix:external-models}.}
\label{tab:main}
\end{table*}

Table~\ref{tab:main} summarises the main result on the reference seed. Counterexample-only SFT lifts greedy Pass@1 modestly above the Qwen3-4B base (0.30 $\to$ 0.35; 95\,\% CI on the paired delta $[-0.030, +0.119]$), entirely consistent with no statistically significant gain at this sample size. Continuing with verifier-grounded RL, regardless of reward shape, yields a substantial further gain to 0.50 [0.43, 0.57] (paired delta versus the base $+0.199$~[+0.13,~+0.27]). Across four seeds the RLVR variants reach 0.49\,$\pm$\,0.02 (dense) and 0.46\,$\pm$\,0.03 (sparse) against an unchanged base of 0.30 (Table~\ref{tab:multiseed}).

Three points are worth emphasising. First, our 4B RLVR model exceeds every 7B math-specialist open-weights model we evaluated (full table in Appendix~\ref{sec:appendix:external-models}); the best 7B baseline (Eurus-2-7B-PRIME at 0.30) is statistically indistinguishable from our untrained 4B base (paired CI $[-0.075, +0.065]$), which isolates the gain to the training method rather than parameter count, and is consistent with the forward/falsification asymmetry reported by \citet{sinha2025refute}. Second, the gain holds in expectation under sampled decoding: RLVR Pass@8 is 0.68, versus 0.57 for the best 7B baseline and 0.49 for the base. Third, our 4B RLVR model falls inside the range of greedy Pass@1 measured for six commercial frontier APIs (0.42--0.58); only one of the six APIs outperforms it. This supports the claim that a 4B model can be made competitive with frontier APIs on this task by training rather than by scale. RLVR also roughly doubles the rate at which the policy negates the conclusion (0.30 $\to$ 0.61); the 0.11 that separates this from Pass@1 is rollouts that negate the conclusion but violate a remaining hypothesis.

\subsection{The imitation trap}
\label{sec:results:trap}

\begin{table}[t]
\centering
\footnotesize
\setlength{\tabcolsep}{3pt}
\begin{tabular}{lcc}
\toprule
Model & {\scriptsize\textsc{SymCE}-prompt} & {\scriptsize\textit{truth-classify}} \\
\midrule
Qwen3-4B base         & 0.27 & 0.36 \\
\quad + SFT           & 0.00 & 0.21 \\
\quad + RLVR (dense)  & 0.34 & 0.73 \\
\quad + RLVR (sparse) & \textbf{0.65} & \textbf{0.84} \\
\bottomrule
\end{tabular}
\caption{Theorem-truth probe (n=347), reference seed 42. Pass@1 (greedy) = fraction of \emph{true} theorems correctly recognised as true, under the answer-extraction rules of \S\ref{sec:experiments:benchmarks}. Four-seed means are in Table~\ref{tab:multiseed}. The base column is conservative: on 62\,\% of items the base states no verdict at all, exhausting the token budget mid-reasoning, and it asserts \texttt{FALSE} on only 2.6\,\%. SFT reverses this, asserting \texttt{FALSE} for 70\,\% of \emph{true} theorems. Four-seed spread for both columns is in Table~\ref{tab:multiseed}.}
\label{tab:trap}
\end{table}

\begin{figure}[t]
\centering
\fbox{\begin{minipage}{0.96\columnwidth}
\footnotesize
\textbf{A true theorem from the theorem-truth probe.}
\emph{Continuous convergence theorem} (Real Analysis): if a sequence of continuous functions satisfies the required convergence assumptions, the limit function is continuous.

\smallskip
\textbf{SFT output.} Counterexample: $X=[0,1]$, $f_n(x)=x^n$, and
\[
f(x)=\begin{cases}0 & x<1,\\ 1 & x=1.\end{cases}
\]
The theorem is true. SFT proposes the classical sequence $f_n(x)=x^n$ anyway, a construction that violates the convergence assumption it would need. RLVR-dense abstains; RLVR-sparse recognises the theorem as true and emits \texttt{NONE}.
\end{minipage}}
\caption{The imitation trap on one item. Appendix~\ref{sec:appendix:qualitative} gives the symmetric sparse-reward failure: a false conjecture the dense model disproves and the sparse model wrongly abstains on.}
\label{fig:qualitative-main}
\end{figure}

Table~\ref{tab:trap} presents what we view as the central scientific finding of this work. Counterexample-only SFT collapses the model's ability to recognise true theorems. Under the counterexample-emit prompt format used throughout training, Pass@1 drops from 0.27 (base) to 0.00 (SFT) on all true theorems, in every seed we ran. The collapse is also visible under an explicit true/false classification prompt (0.36 $\to$ 0.21), and there it takes a particularly stark form: where the base model mostly declines to commit, the supervised-only model actively asserts that 70\,\% of true theorems are false. The base model retains genuine calibration; the supervised-only model does not. Figure~\ref{fig:qualitative-main} shows one such item.

Continuing from the same SFT adapter, verifier-grounded RL with a \textit{sparse} outcome-only reward repairs and exceeds the base under both prompt styles (0.65 and 0.84 respectively). The dense reward also repairs the more demanding \textsc{SymCE}-prompt probe, but only to just above the base (0.34 against 0.27), where sparse reaches 0.65. The dense and sparse variants are statistically indistinguishable on the in-distribution counterexample benchmark of \S\ref{sec:results:main} (Table~\ref{tab:main}) but differ by 31 percentage points on this probe on the reference seed, and by 33\,$\pm$\,8 points across four seeds. Appendix~\ref{sec:appendix:memorisation} shows the dissociation is larger still on the subset of true theorems whose mutations were never seen in training, so it is not a memorisation artefact.

This dissociation, equal in-distribution Pass@1 with a large out-of-distribution gap, is interpretable as evidence that the two reward shapes induce qualitatively different policies despite producing the same counterexample-generation performance. We give the mechanism in \S\ref{sec:results:reason}.

\subsection{Why it happens: a reason-taxonomy analysis}
\label{sec:results:reason}

\begin{figure}[t]
\centering
\includegraphics[width=\columnwidth]{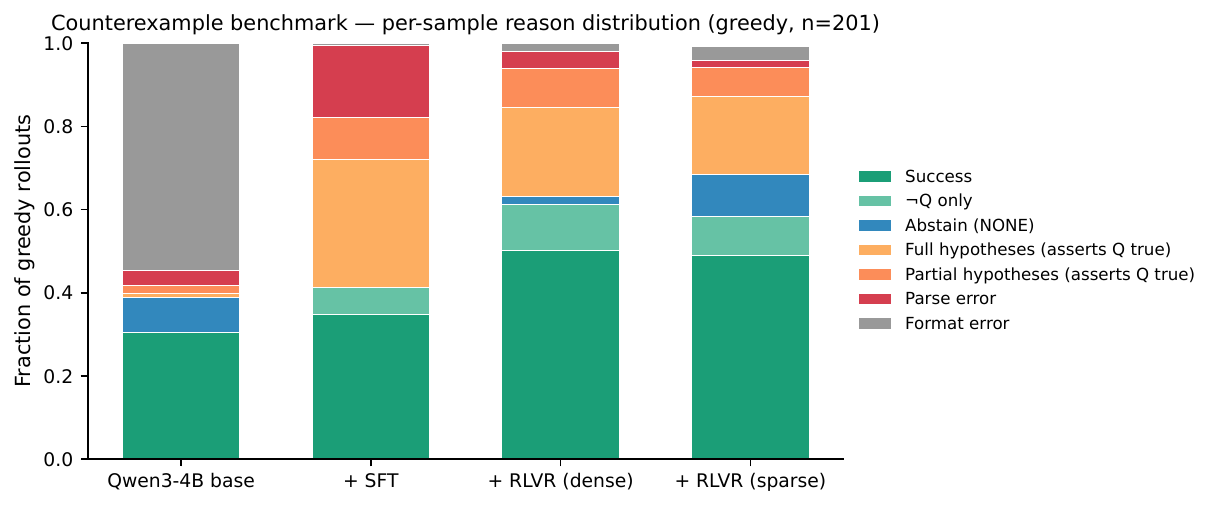}
\caption{Per-sample outcome distribution on the counterexample benchmark (greedy, n=201). The oracle assigns each rollout exactly one outcome: \emph{success} ($q_{\neg}{=}1$ and all hypotheses satisfied), \emph{$\neg$Q only} ($q_{\neg}{=}1$ but not all hypotheses), \emph{abstain} (the \texttt{NONE} sentinel on a false conjecture), \emph{hyp-only} (all hypotheses satisfied but $q_{\neg}{=}0$, so the policy asserts the false conjecture is true), \emph{partial}, \emph{parse error} and \emph{no tag}. The released records carry these as the codes \texttt{success}, \texttt{neg\_q\_only}, \texttt{none\_on\_false}, \texttt{hyp\_only}, \texttt{partial}, \texttt{oracle\_parse\_error} and \texttt{no\_tag}.}
\label{fig:reason}
\end{figure}

We exploit the structured reward to label every test rollout with one of seven mutually exclusive \textit{outcomes} (Figure~\ref{fig:reason}). Three observations explain the imitation trap.

\textbf{SFT trades format errors for logical errors.} The base model rarely emits the trained format at all: 55\,\% of its rollouts carry no tag. SFT solves this, but the corresponding mass does not move uniformly into success; instead, the hyp-only category, in which the policy satisfies the remaining hypotheses but asserts the false conclusion is true, rises from 1\,\% to 31\,\%. The supervised-only model learned the \emph{form} of a counterexample without learning the calibration: it confidently emits objects that fail the negated-conclusion check.

\textbf{Sparse RL re-learns abstention.} The abstention rate is 0\,\% under SFT, 2\,\% under dense RL, and 10\,\% under sparse RL. The sparse reward yields equal score for emitting a wrong witness and emitting \texttt{NONE}, so any rollout the policy is uncertain about is funnelled into abstention; under the dense reward, the partial-credit term $\alpha \cdot (s_{\text{hyp}} / n_{\text{hyp}})$ makes ``try and partially succeed'' strictly better than ``abstain''. This is the causal mechanism for the 31-percentage-point gap of Table~\ref{tab:trap}.

\textbf{Identical success rates hide diverging failure modes.} While dense and sparse RL achieve nearly identical success fractions (0.502 vs 0.497), dense rewards push the remaining probability mass into over-eager hyp-only errors, whereas sparse rewards push it into abstention.

We conclude that when a verifier yields a structured boolean tuple, an outcome-only aggregator preserves calibration better than partial credit. Rewarding satisfaction of some but not all hypotheses actively trains over-eager counterexample emission. The claim is specific to that term rather than to dense rewards in general, and our dense reward is not a strawman: its conclusion weighting ($\beta{=}0.65 > \alpha{=}0.25$) provably precludes the naive hypothesis-only dominance failure (\S\ref{sec:method:reward}), and it was selected after exploring several dense variants.

\subsection{Cross-domain consistency}
\label{sec:results:domain}

\begin{figure}[t]
\centering
\includegraphics[width=0.98\columnwidth]{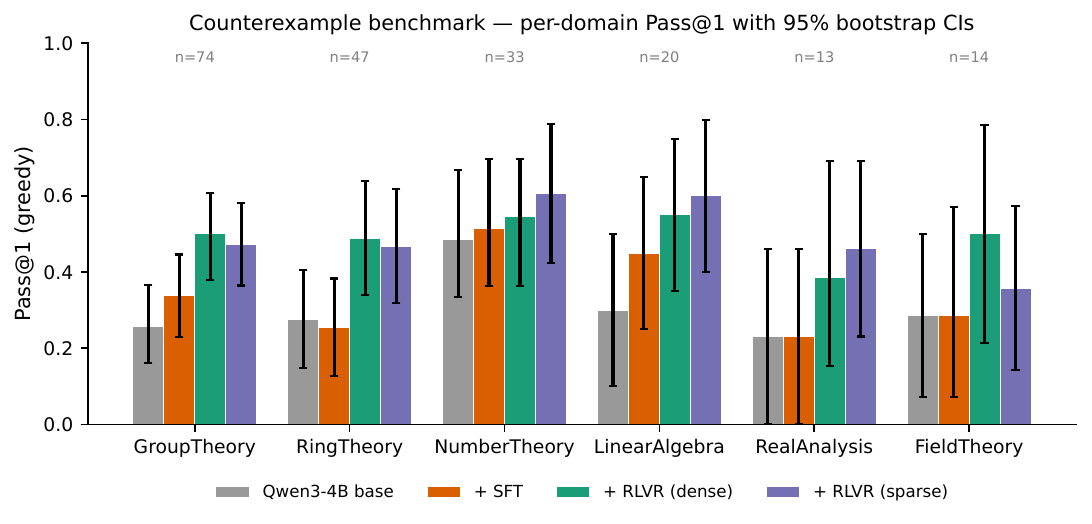}
\caption{Per-domain counterexample Pass@1 (greedy) with 95\,\% bootstrap CIs. Annotations on top: number of test items per domain. The largest \textit{absolute} gain is on LinearAlgebra (+0.30); the most striking result is on RealAnalysis, where RLVR-sparse gains +0.23 absolute despite the domain being only 6.7\,\% of the RLVR training distribution.}
\label{fig:domain}
\end{figure}

Figure~\ref{fig:domain} stratifies the main result by domain. Both RLVR variants improve on the base in every domain, with no sign of domain collapse or narrow specialisation. The largest relative gain is on Real Analysis (0.23 $\to$ 0.46), despite it being the smallest slice of the RLVR training set at 6.7\,\%. This is the inverse of what an over-fitting model would do, and is consistent with the policy learning a transferable skill rather than a domain-conditional shortcut.

Dense and sparse 95\,\% CIs overlap in almost every domain, supporting the in-distribution equivalence of \S\ref{sec:results:main} per-domain rather than as an aggregation artefact.

\subsection{Out-of-domain transfer}
\label{sec:results:ood}

\begin{table}[t]
\centering
\small
\setlength{\tabcolsep}{4pt}
\begin{tabular}{lcccc}
\toprule
Benchmark (n) & base & SFT & dense & sparse \\
\midrule
GSM8K (1319)       & 0.83 & 0.82 \tiny{$\pm$.03} & 0.88 \tiny{$\pm$.02} & \textbf{0.88} \tiny{$\pm$.01} \\
MATH-500 (500)     & 0.13 & 0.55 \tiny{$\pm$.01} & \textbf{0.65} \tiny{$\pm$.02} & 0.65 \tiny{$\pm$.01} \\
MMLU-CM (100)      & 0.34 & 0.54 \tiny{$\pm$.03} & 0.70 \tiny{$\pm$.05} & \textbf{0.72} \tiny{$\pm$.04} \\
AIME-2024 (30)     & 0.00 & 0.07 \tiny{$\pm$.03} & \textbf{0.08} \tiny{$\pm$.03} & 0.08 \tiny{$\pm$.02} \\
\bottomrule
\end{tabular}
\caption{Out-of-domain transfer under standard chain-of-thought prompting (greedy, Pass@1), mean\,$\pm$\,sd over all four training seeds. Identical prompt template across all four models; our SFT/RLVR pipeline contains no examples or answer-format training from these benchmarks.}
\label{tab:ood}
\end{table}

Table~\ref{tab:ood} measures transfer to four standard mathematical-reasoning benchmarks. Training improves on the base everywhere, with sparse and dense comparable. All four models are evaluated under identical chain-of-thought prompting, and our SFT and RLVR data contain no GSM8K, MATH, MMLU or AIME examples and no training on their answer-extraction formats, so the improvements are attributable to transfer rather than to format adaptation in the training pipeline.

Part of the MATH-500 jump is nonetheless format adaptation: the base scores 0.13 largely because it does not follow the answer-extraction template, which SFT installs. GSM8K isolates the reasoning gain, because there the base already formats answers correctly (0.83), yet RLVR still reaches 0.88 under the identical prompt while SFT alone does not improve on the base at all (0.82\,$\pm$\,0.03). The transfer therefore arrives with the reinforcement stage, not the imitation stage.

We interpret these results as supporting the broader narrative motivated in the introduction: counterexample-style supervision is a fundamental component of mathematical reasoning that conventional pre-training under-installs, rather than a narrow side-skill.

\subsection{Robustness across seeds and model families}
\label{sec:results:robustness}

\begin{table}[t]
\centering
\small
\setlength{\tabcolsep}{4pt}
\begin{tabular}{lcccc}
\toprule
Metric & base & SFT & dense & sparse \\
\midrule
\multicolumn{5}{l}{\textit{Qwen3-4B, mean\,$\pm$\,sd over 4 seeds}} \\
\midrule
\textsc{SymCE} Pass@1  & 0.30 & 0.34 \tiny{$\pm$.02} & \textbf{0.49} \tiny{$\pm$.02} & 0.46 \tiny{$\pm$.03} \\
\textsc{SymCE} Pass@8  & 0.49 & 0.62 \tiny{$\pm$.02} & \textbf{0.66} \tiny{$\pm$.02} & 0.64 \tiny{$\pm$.02} \\
Truth \textsc{SymCE}-prompt & 0.27 & 0.00 \tiny{$\pm$.00} & 0.33 \tiny{$\pm$.04} & \textbf{0.66} \tiny{$\pm$.04} \\
Truth-classify & 0.36 & 0.16 \tiny{$\pm$.05} & 0.71 \tiny{$\pm$.09} & \textbf{0.75} \tiny{$\pm$.13} \\
\midrule
\multicolumn{5}{l}{\textit{Gemma-3-4B, seed 56}} \\
\midrule
\textsc{SymCE} Pass@1  & 0.21 & 0.22 & \textbf{0.39} & 0.34 \\
Truth-classify & 0.49 & 0.01 & 0.53 & \textbf{0.59} \\
\bottomrule
\end{tabular}
\caption{Robustness of the headline findings. \textsc{SymCE} Pass@$k$ is the counterexample benchmark (n=201); the two Truth rows are the theorem-truth probe (n=347) under the two prompt styles of \S\ref{sec:experiments:benchmarks}. Qwen3-4B seeds are 42, 19, 31 and 64; the base is training-seed independent. Bold marks the best value in a row.}
\label{tab:multiseed}
\end{table}

Table~\ref{tab:multiseed} repeats the pipeline with three further seeds and on a second model family; per-seed values are in Appendix~\ref{sec:appendix:seeds}. The imitation trap is not a seed artefact: counterexample-only SFT scores exactly 0.00 on the \textsc{SymCE}-prompt probe in all four Qwen seeds. The sparse-over-dense calibration gap on that probe is positive in every seed, at $+0.33 \pm 0.08$. In-domain success stays within noise between the two reward shapes, dense ahead by 0.02 on average, consistent in sign but far inside the test-set bootstrap width. And the phenomenon crosses model families: on Gemma-3-4B the collapse is more total still, from 0.49 to 0.01 under explicit classification and from 0.40 to 0.00 on the \textsc{SymCE}-prompt probe, with both RLVR variants restoring classification above the base. It replicates on a second Gemma seed (24): the same two probes fall from 0.49 to 0.00 and from 0.40 to 0.00, and dense RLVR restores classification to 0.54.

On \textit{truth-classify} the sparse advantage is smaller and less consistent: sparse leads dense by $0.04 \pm 0.15$, positive in three seeds and negative in seed 64. We therefore restrict the sparse-over-dense calibration claim to the probe that shares the training prompt style, where it holds in every seed. Appendix~\ref{sec:appendix:qualitative} gives worked examples of both failure modes.

\section{Discussion}
\label{sec:discussion}

\paragraph{This is not generic catastrophic forgetting.} A natural reading of \S\ref{sec:results:trap} is that counterexample-only SFT simply induces forgetting, and that standard mitigations such as replay are the appropriate control. Four properties argue otherwise. The failure is \textit{selective}: after SFT, general mathematical competence improves substantially (MATH-500 0.13 $\to$ 0.55, Table~\ref{tab:ood}) while true-theorem recognition falls to 0.00, which is closer to the hallucination tax \citep{ouyang2025treecut, song2025hallucination} than to forgetting. The repair \textit{exceeds} the baseline (0.27 $\to$ 0.66), which recovering previously seen behaviour does not predict. It occurs with \textit{zero true theorems in training}, purely by changing the reward's treatment of abstention; a replay baseline would alter that distribution and dissolve the phenomenon, so the controlled comparison is dense versus sparse, which holds data and initialisation fixed and still differs by 33 points. Finally the mechanism is \textit{localised} to a specific behavioural shift (\S\ref{sec:results:reason}), and generic forgetting offers no account of why two reward shapes on the same checkpoint should differ at all.

\paragraph{Why sparse generalises better on calibration.} The mechanism in \S\ref{sec:results:reason} is interpretable. Under the dense reward, the partial-credit term $\alpha \cdot (s_{\text{hyp}} / n_{\text{hyp}})$ guarantees a positive reward whenever the policy satisfies at least one remaining hypothesis, even when it asserts the (false) conclusion is true. This actively trains the policy to \textit{always emit a counterexample expression}: doing so on a false conjecture yields up to $\alpha + \gamma = 0.35$ and never less than $\gamma = 0.10$ when the witness parses, much better than the $-0.20$ of abstention. Sparse rewards remove this incentive: the policy is rewarded only for correctly negating the conclusion, and abstention is no worse than a wrong witness. The behavioural consequence is the higher abstention rate on the in-distribution test set (10\,\% vs 2\,\%) and, crucially, the preservation of abstention as a viable action when the policy is shown a \textit{true} theorem at test time. The dense reward does not preserve this; the partial-credit term is in tension with calibration.

We conjecture this generalises to other settings where (i) the reward signal already returns a structured boolean tuple, and (ii) the failure mode of interest is the policy answering when it should abstain, rather than failing to produce well-formed output. Whether this conjecture survives at scale is left to future work.

\paragraph{Deterministic verification beats LLM-as-judge where the verification problem is decidable.} The verifier substrate is reproducible, fast (median 132~ms), audited end-to-end at 97.7\,\% decision accuracy over 177 human-checked verifier decisions, and immune to judge-model drift. We would add that this shifts, rather than removes, the dependence on a language model: an LLM-as-judge protocol incurs stochastic error at every evaluation, whereas construction-time LLM dependence is frozen once and can be audited once, which is what \S\ref{sec:method:audit} does. Its scope is bounded by what is decidable in Python with SymPy/Z3; for the algebraic subdomains in this paper this is sufficient. Where the verification problem is decidable a deterministic verifier is the stronger choice; LLM-judging, as used by COUNTERMATH \citep{li2025countermath}, remains necessary for domains that fall outside it.

\paragraph{Counterexample reasoning as a fundamental capability.} The cross-domain consistency of \S\ref{sec:results:domain} and the out-of-domain transfer of \S\ref{sec:results:ood} together support a stronger reading of the contribution: training on structured falsification improves general mathematical reasoning, not only disproof. The Real Analysis result (6.7\,\% training share, $+$0.23 absolute gain) and the $+$52 percentage-point gain on MATH-500 under unchanged prompting are the strongest evidence for this reading. The literature on the ``hallucination tax'' \citep{song2025hallucination} predicts that standard imitation-only training degrades calibration; the present results suggest that verifier-grounded RL with an outcome-only reward avoids that pathology and installs a transferable inverse-reasoning skill.

\section{Conclusion}
\label{sec:conclusion}

We have presented \textsc{SymCE}, a deterministic, large-scale dataset and verification environment for counterexample generation on undergraduate algebra and real analysis, and used it to identify and explain an \emph{imitation trap}: supervised fine-tuning on counterexample-only data collapses the policy's ability to recognise true theorems, and verifier-grounded reinforcement learning with a sparse outcome-only reward repairs and exceeds the original calibration where a partial-credit dense reward does not. The collapse reproduces in every training seed and on a second model family.
A Qwen3-4B model trained this way exceeds every 7B open-weights math specialist we evaluated on the in-domain benchmark, remains competitive with frontier commercial APIs, and transfers under unchanged prompting to GSM8K, MATH-500 and MMLU-college-math, evidence that counterexample reasoning is a fundamental, under-installed skill rather than a narrow specialisation.
We release the dataset, the verifier modules, the human-audit annotations, the witness-schema generator, and the full evaluation pipeline, and we view extension to topology and higher-order logic, broader mutation strategies, larger scales and alternative RL algorithms, and integration with a Lean back-end as the natural next steps.

\section*{Limitations}
\label{sec:limitations}

\paragraph{Scale, parameterisation and RL algorithm.} Every trained variant here is a 4B model adapted with LoRA and optimised with GRPO, so three axes remain untested: substantially larger scales, full-parameter fine-tuning, and alternative RL algorithms. Full fine-tuning with GRPO rollouts was not feasible on our compute (a single RTX~5090, 32\,GB), so we cannot separate properties of the environment from properties of GRPO. Two observations mitigate the concern that the findings are adapter artefacts: they replicate across four seeds and a second model family under the same adapter configuration, including an exact 0.00 collapse in all six SFT runs, and the mechanism we identify operates at the level of the optimisation objective rather than the parameterisation. The scaling question is the most important one we leave open.

\paragraph{Verifier scope.} Our substrate is Python with SymPy and Z3. This suffices for the six subdomains we evaluate but not for topology, general higher-order logic, or claims whose verification reduces to undecidable fragments. A Lean backend, or continuous-FOL extensions for metric spaces, is a natural follow-up.

\paragraph{Audit scope.} The 177-decision study of \S\ref{sec:method:audit} was annotated by a single author, so we report no inter-annotator agreement, and 177 decisions bound the pipeline's error rate only coarsely. A larger, independently annotated audit would support stronger claims about the full verifier collection.

\paragraph{Mutation strategy and scope of the resource claim.} We mutate only by dropping a single hypothesis. This was deliberate: it gives every false conjecture a known, attributable cause of falsity, which is what makes the per-hypothesis boolean interface and the outcome taxonomy possible. Conclusion mutation, multi-hypothesis dropping and AST-level mutation \citep{wu2024alchemy} would enrich the distribution but complicate attribution, and remain unexplored. \textsc{SymCE} should accordingly be read as a deterministic falsification environment rather than a benchmark spanning the full space of mathematical fallacies.

\paragraph{No COUNTERMATH cross-evaluation.} Cross-evaluation on COUNTERMATH \citep{li2025countermath} would compare our models against the published frontier on a benchmark with broader domain coverage. The authors do not release the test labels or evaluation scripts, and predicted outputs must be submitted for LLM-judged scoring on their side, which diverges from this paper's focus on open, deterministic verification. We leave this for future work.

\paragraph{Teacher model dependence.} The verifiers are generated by a single teacher model (Gemini 3.1 Pro) and the witness schemas by a second (DeepSeek). Although the audit found no verifier-logic errors, the pipeline retains a dependence on teacher output quality. Multi-model consensus generation, in which several teachers must agree on a verifier before it enters the corpus, is a valuable extension we have not implemented.

\section*{Ethics and Risk Considerations}
The work studies counterexample generation for undergraduate-level algebraic and real-analysis theorems, using LLM-generated Python verifiers and a hypothesis-dropping mutation pipeline. The artefacts produced (\textsc{SymCE}, a dataset of false mathematical conjectures, per-theorem verifier modules, and a small 4B model specialised for falsifying such conjectures) do not present obvious dual-use risks beyond those general to research on LLM mathematical reasoning. Per-artefact licences are listed in Appendix~\ref{sec:appendix:licences}. The dataset contains no human-derived data, no personal information, and no naturally occurring text from which PII could be derived; the trained models are released for research use and are not deployed in any user-facing setting; and the methodology does not enable capabilities that meaningfully extend beyond existing open-source mathematical reasoning models.

\section*{Acknowledgements}
\label{sec:acks}

We gratefully acknowledge \textbf{Sukai Labs} and its CEO, Ahmed Sedouki, for believing in this research and providing access to the NVIDIA RTX~5090 (32\,GB) GPU on which all training and evaluation runs in this paper were performed. We also thank the ACL ARR reviewers and the action editor, whose requests for multi-seed replication, a second model family, and a stronger verifier audit shaped the evidence reported here.

\bibliography{custom}

\appendix

\section{Full Hyperparameters}
\label{sec:appendix:hyperparameters}

Table~\ref{tab:hyperparams} lists the exact hyperparameters used for the SFT and RLVR runs reported in this paper. The reference-seed runs were launched from \texttt{configs/full\_run.yaml} (SFT), \texttt{configs/full\_run\_dense\_test.yaml} (dense RLVR) and \texttt{configs/full\_run\_sparse.yaml} (sparse RLVR); the additional seeds are in \texttt{configs/seeds/}. Each run directory also carries the fully resolved \texttt{config.json} it was launched with, which is the authoritative record.

\begin{figure}[!h]
\centering
\includegraphics[width=\columnwidth]{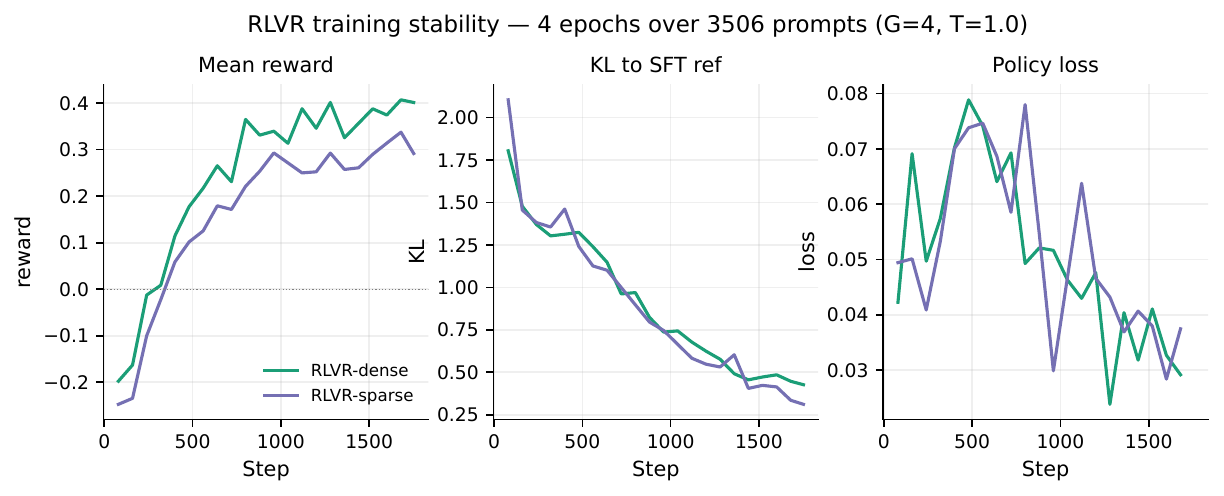}
\caption{Per-step training curves for the dense and sparse RLVR runs at the reference seed ($\approx$\,1{,}752 steps over four epochs of 3{,}506 prompts). Mean reward rises monotonically and crosses zero by step $\sim$240 for both variants; KL to the SFT reference decays smoothly; loss is bounded and oscillatory but stable.}
\label{fig:training}
\end{figure}

Figure~\ref{fig:training} shows the per-step training curves for both RLVR runs. The reward trajectory rises from approximately $-0.20$ at step 80 to $+0.40$ (dense) / $+0.30$ (sparse) at convergence; the KL penalty to the SFT reference declines from $\sim 2.0$ to $\sim 0.5$ as the policy explores away from imitation; the policy loss is bounded and oscillatory but does not diverge. Both runs complete the full four epochs without any restart, gradient explosion or reward collapse.

\begin{table}[!h]
\centering
\footnotesize
\setlength{\tabcolsep}{4pt}
\begin{tabular}{lr}
\toprule
\multicolumn{2}{l}{\textbf{Supervised fine-tuning (SFT)}} \\
\midrule
Base model & \texttt{unsloth/Qwen3-4B} \\
Chat template & \texttt{qwen-3} \\
Max sequence length & 2048 \\
Precision & bfloat16 \\
LoRA rank $r$ / scale $\alpha$ & 16 / 16 \\
LoRA dropout & 0 \\
LoRA targets & all attn.\ + MLP \\
Gradient checkpointing & unsloth \\
Per-device batch & 16 \\
Gradient accumulation & 2 \\
Effective batch & 32 \\
Epochs & 3 \\
Learning rate & $2 \times 10^{-4}$ \\
LR schedule / warmup & cosine / 0.05 \\
Optimiser & AdamW \\
Weight decay & 0.0 \\
Max grad norm & 1.0 \\
Eval-loss split fraction & 0.05 \\
Seed & 42 \\
\midrule
\multicolumn{2}{l}{\textbf{RLVR (GRPO from SFT adapter)}} \\
\midrule
Algorithm & TRL \texttt{GRPOTrainer} \\
Rollout backend & vLLM (\texttt{gpu\_util}\,=\,0.5) \\
Max LoRA rank & 16 \\
$G$ (generations / prompt) & 4 \\
Max prompt / completion tok. & 768 / 1024 \\
Temperature & 1.0 \\
Top-$p$ & 1.0 \\
Per-device batch & 4 \\
Gradient accumulation & 8 \\
Effective batch & 32 \\
Epochs & 4 \\
Total steps & $\approx$\,1{,}752 \\
Learning rate & $5 \times 10^{-6}$ \\
LR schedule / warmup & constant / 0.1 \\
PPO clip $\varepsilon$ & 0.2 \\
KL coefficient $\beta_{\text{KL}}$ & 0.02 \\
Max grad norm & 0.5 \\
Oracle timeout & 8 s \\
Reward-pool workers & 4 \\
Seed & 42 \\
\midrule
\multicolumn{2}{l}{\textbf{Oracle reward weights} (Eq.~\ref{eq:reward})} \\
\midrule
\multicolumn{2}{l}{Dense $(\alpha,\beta,\gamma,\delta,\varepsilon)=(0.25,0.65,0.10,0.30,1.00)$} \\
\multicolumn{2}{l}{Sparse $(\alpha,\beta,\gamma,\delta,\varepsilon)=(0,1,0,0,1)$} \\
\bottomrule
\end{tabular}
\caption{Hyperparameters used for SFT and both RLVR variants, as recorded by the reference-seed runs. The one setting that is not uniform is the SFT epoch count: seed 42 uses 3 epochs, and seeds 19, 31 and 64 use 2, after we checked that the two schedules performed equivalently and preferred the shorter one. SFT wall-clock on a single RTX~5090 32\,GB is under five minutes; each RLVR run is approximately nine hours.}
\label{tab:hyperparams}
\end{table}

\section{Human Audit of Verifier Decisions}
\label{sec:appendix:audit}

We sampled 177 model-generated witnesses from the held-out test split, drawn across four seeds and three training configurations (dense 85, sparse 61, SFT 31) and all six subdomains, and split by the verifier's own verdict: 101 that it accepted and 76 that it rejected. An author with graduate-level training in the relevant areas judged, independently of the verifier, whether each witness is genuinely a valid counterexample. Table~\ref{tab:audit} reports the composition.

\begin{table}[h]
\centering
\small
\begin{tabular}{lrrr}
\toprule
Domain & accepted & rejected & total \\
\midrule
GroupTheory   & 31 & 27 & 58 \\
RingTheory    & 20 & 15 & 35 \\
NumberTheory  & 23 &  8 & 31 \\
LinearAlgebra & 11 & 11 & 22 \\
FieldTheory   &  8 &  8 & 16 \\
RealAnalysis  &  8 &  7 & 15 \\
\midrule
\textbf{Total} & \textbf{101} & \textbf{76} & \textbf{177} \\
\bottomrule
\end{tabular}
\caption{Composition of the human audit, split by the verifier's verdict. Decision accuracy: 99/101 accepted witnesses confirmed valid, 74/76 rejected witnesses confirmed invalid, 173/177 = 97.7\,\% overall. Annotations are released at \texttt{human\_eval/}.}
\label{tab:audit}
\end{table}

\paragraph{The four disputed decisions.} Two verifier acceptances were judged invalid: one witness satisfied the remaining hypotheses without actually negating the conclusion, and one supplied a malformed field construction. Two rejections were judged wrong, and in neither case is the verifier module at fault. In \texttt{bw\_seq\_rn\_\_h3} the emitted witness was valid but the witness-schema description showed a bare \texttt{Matrix} where the verifier's namespace requires \texttt{sympy.Matrix}; the fault is in the schema text, not the verifier logic. In \texttt{word\_problem\_direct\_product} the conjecture remained true after the hypothesis drop, and the model correctly emitted \texttt{NONE}; the verifier ``rejected'' only because a counterexample was expected where none exists. This is the single case in 177 that we attribute to hypothesis-drop label noise, and the single case attributable to schema quality.

Among the confirmed-correct rejections the annotator noted several where the mathematics was right and only the Python namespace or API usage was wrong, such as \texttt{Matrix} for \texttt{sympy.Matrix}; only 9 of the 76 rejections carry free-text annotations, so we cannot estimate a rate.

\section{Answer Extraction and the Scoring Correction}
\label{sec:appendix:scoring}

\S\ref{sec:experiments:benchmarks} adopts strict answer-extraction rules for the two calibration probes. This appendix reports both readings so the effect is auditable.

The evaluation harness that produced our raw records used, as a last resort when no answer frame was present, a parser matching any bare occurrence of \emph{true} or \emph{false} anywhere in the completion, taking the last one. On completions that exhaust the generation budget mid-reasoning this converts unfinished chains of thought into apparent verdicts. The effect is confined almost entirely to the untrained base model, which reasons at length and frequently fails to commit within budget; the trained models emit the required format and are barely affected.

\begin{table}[h]
\centering
\footnotesize
\setlength{\tabcolsep}{3pt}
\begin{tabular}{lcccc}
\toprule
& \multicolumn{2}{c}{\scriptsize\textsc{SymCE}-prompt} & \multicolumn{2}{c}{\scriptsize\textit{truth-classify}} \\
\cmidrule(lr){2-3}\cmidrule(lr){4-5}
Model & strict & permissive & strict & permissive \\
\midrule
Qwen3-4B base        & 0.27 & 0.86 & 0.36 & 0.71 \\
\quad + SFT          & 0.00 & 0.00 & 0.21 & 0.23 \\
\quad + RLVR (dense) & 0.34 & 0.37 & 0.73 & 0.76 \\
\quad + RLVR (sparse)& 0.65 & 0.69 & 0.84 & 0.84 \\
\bottomrule
\end{tabular}
\caption{Theorem-truth probe under both answer-extraction rules (reference seed 42, n=347). ``Permissive'' reproduces what the original evaluation harness recorded; the paper reports the strict column throughout.}
\label{tab:scoring}
\end{table}

Two properties of the finding are invariant to the choice. Counterexample-only SFT scores exactly 0.00 on the \textsc{SymCE}-prompt probe under both rules and in every seed, and the sparse-over-dense gap on that probe is 0.31 under both. What the choice does affect is the base model's apparent calibration, and hence whether RLVR is described as exceeding it.

We validated that the strict \textit{truth-classify} rule does not under-credit models that do answer. Across every cell in our corpus, each response the rule records as stating no verdict is an unfinished completion; no completion that terminated normally is scored as having failed to answer. The accepted answer frames were derived by inspecting the completions that the narrow \texttt{Answer:\ TRUE} pattern alone would have missed, and cover the phrasings both model families actually produce, including \texttt{\textbackslash boxed\{TRUE\}} and ``the claim is \textbf{TRUE}''.

\section{Per-Seed Results}
\label{sec:appendix:seeds}

Table~\ref{tab:perseed} gives the individual seed values behind the means of Table~\ref{tab:multiseed}. The base model is training-seed independent and is evaluated once per suite.

\begin{table}[h]
\centering
\small
\setlength{\tabcolsep}{4pt}
\begin{tabular}{llcccc}
\toprule
Metric & Arm & 42 & 19 & 31 & 64 \\
\midrule
\multirow{3}{*}{\textsc{SymCE} Pass@1}
 & SFT    & 0.348 & 0.358 & 0.323 & 0.338 \\
 & dense  & 0.502 & 0.478 & 0.458 & 0.502 \\
 & sparse & 0.497 & 0.438 & 0.463 & 0.453 \\
\midrule
\multirow{3}{*}{\textsc{SymCE} Pass@8}
 & SFT    & 0.637 & 0.617 & 0.592 & 0.632 \\
 & dense  & 0.677 & 0.632 & 0.672 & 0.662 \\
 & sparse & 0.672 & 0.622 & 0.632 & 0.637 \\
\midrule
\multirow{3}{*}{Truth \textsc{SymCE}-prompt}
 & SFT    & 0.000 & 0.000 & 0.000 & 0.000 \\
 & dense  & 0.343 & 0.294 & 0.308 & 0.386 \\
 & sparse & 0.654 & 0.715 & 0.660 & 0.608 \\
\midrule
\multirow{3}{*}{Truth-classify}
 & SFT    & 0.210 & 0.101 & 0.173 & 0.161 \\
 & dense  & 0.729 & 0.801 & 0.594 & 0.700 \\
 & sparse & 0.839 & 0.813 & 0.787 & 0.550 \\
\bottomrule
\end{tabular}
\caption{Per-seed values. The sparse-over-dense gap on the \textsc{SymCE}-prompt probe is positive in every seed ($+0.31$, $+0.42$, $+0.35$, $+0.22$); on \textit{truth-classify} it is not ($+0.11$, $+0.01$, $+0.19$, $-0.15$), which is why \S\ref{sec:results:robustness} restricts the claim to the \textsc{SymCE}-prompt probe.}
\label{tab:perseed}
\end{table}

\section{Theorem-Truth Robustness: Seen vs Unseen Subsets}
\label{sec:appendix:memorisation}

The 347-item theorem-truth probe partitions into true theorems whose \emph{mutated} form appeared in the RLVR training corpus (\texttt{extra\_stratified}, n=150) and those whose mutated form did not (\texttt{from\_test\_set}, n=197). Because both subsets are drawn from the same theorem source, comparing them bounds how much of the trained models' calibration is memorisation.

\begin{table}[h]
\centering
\small
\begin{tabular}{lccc}
\toprule
Model & unseen & seen & $\Delta$ \\
\midrule
Qwen3-4B base       & 0.244 & 0.307 & $+0.063$ \\
\quad + SFT         & 0.000 & 0.000 & $\phantom{+}0.000$ \\
\quad + RLVR dense  & 0.274 & 0.433 & $+0.159$ \\
\quad + RLVR sparse & 0.614 & 0.707 & $+0.092$ \\
\bottomrule
\end{tabular}
\caption{\textsc{SymCE}-prompt Pass@1 by subset (greedy, strict scoring, reference seed 42). $\Delta$ = seen $-$ unseen.}
\label{tab:memorisation}
\end{table}

Two readings. The base model shows a small positive $\Delta$ ($+0.06$), which sets the scale of any residual difficulty difference between the subsets rather than of memorisation, since the base saw neither. The trained models show larger gaps ($+0.09$ to $+0.16$), consistent with some memorisation of theorem identity. Crucially, the sparse-over-dense gap on the \emph{unseen} subset alone is $0.614 - 0.274 = +0.34$, larger than on the full probe, so the central dissociation of this paper is not a memorisation artefact.

\section{Qualitative Examples}
\label{sec:appendix:qualitative}

Figure~\ref{fig:qualitative-main} shows the imitation trap on a true theorem: the SFT model emits a syntactically valid, well-formatted counterexample expression that none the less fails the negated-conclusion check, which is a hyp-only failure in the taxonomy of \S\ref{sec:results:reason}. Figure~\ref{fig:qualitative} below shows the symmetric failure mode of the sparse reward.

\begin{figure}[h]
\centering
\fbox{\begin{minipage}{0.94\columnwidth}
\footnotesize
\textbf{A false conjecture from the counterexample benchmark.}
\emph{Minimal polynomial of $\sqrt{a}+\sqrt{b}$} (Field Theory): the original theorem requires that $a$, $b$ and $ab$ are all non-squares in the base field. The benchmark removes the assumption that $a$ is not a square.

\smallskip
\textbf{RLVR-dense witness.} $(F,a,b) = (\mathbb{Q},1,2)$. Since $1$ is a square in $\mathbb{Q}$, the claimed quartic is no longer the minimal polynomial of $1+\sqrt{2}$:
\[
x^2 - 2x - 1 \;\neq\; x^4 - 6x^2 + 1 .
\]
RLVR-dense finds this compact counterexample; RLVR-sparse abstains and emits \texttt{NONE}.
\end{minipage}}
\caption{A sparse-reward failure: improved calibration on true theorems costs false abstentions on genuinely false conjectures. We discuss the trade-off in \S\ref{sec:discussion}.}
\label{fig:qualitative}
\end{figure}

\section{Per-Model Baseline Breakdowns}
\label{sec:appendix:external-models}

\begin{figure}[h]
\centering
\includegraphics[width=\columnwidth]{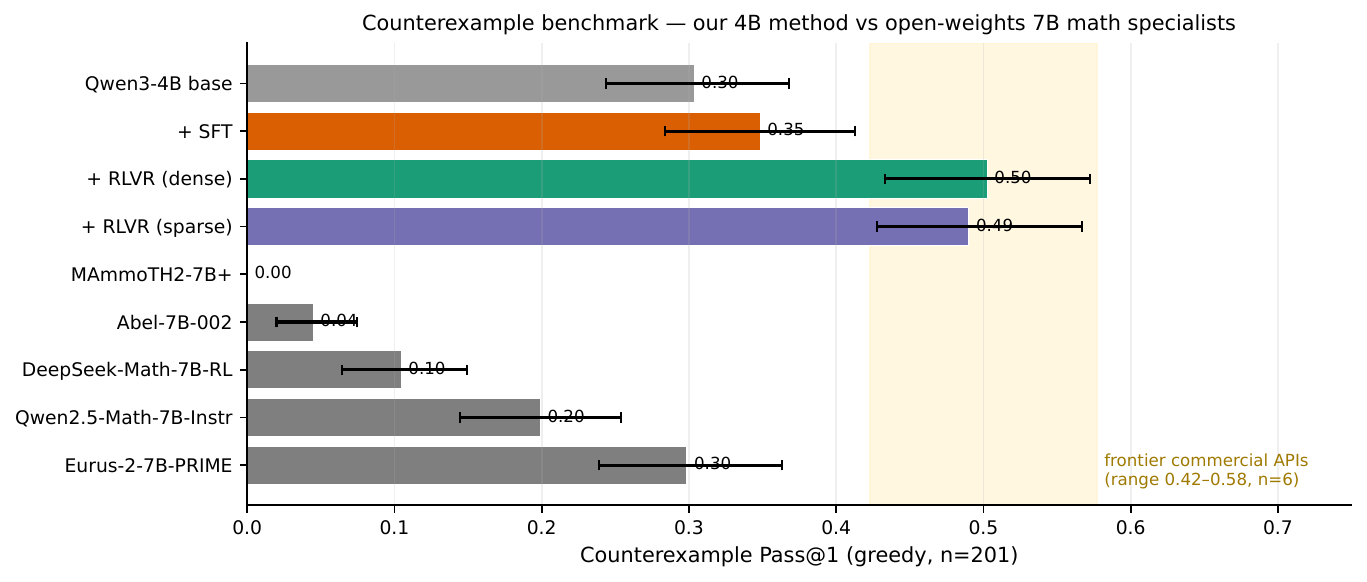}
\caption{Counterexample Pass@1 (greedy) for all comparators evaluated. Our trained 4B models are in colour; open-weights 7B specialists are in grey; the shaded yellow band shows the range across six frontier commercial APIs.}
\label{fig:baselines}
\end{figure}

Table~\ref{tab:oss-baselines} reports the per-model counterexample-benchmark scores for the five 7B open-weights math specialists summarised in the body. Table~\ref{tab:api-baselines} reports the same for the six commercial-API frontier models that contribute to the ``frontier API range'' band in Table~\ref{tab:main}. We additionally report the Qwen2.5-Math-1.5B base in Table~\ref{tab:oss-baselines} as a smaller comparison point. Figure~\ref{fig:baselines} shows all comparators together.

\begin{table*}[h]
\centering
\small
\begin{tabular}{lcccc}
\toprule
Model & Pass@1 (greedy) & Pass@1 (sampled) & Pass@3 (sampled) & Pass@8 (sampled) \\
\midrule
Eurus-2-7B-PRIME            & 0.299 \tiny{[0.24, 0.36]} & 0.292 \tiny{[0.24, 0.34]} & 0.458 \tiny{[0.40, 0.52]} & 0.567 \tiny{[0.50, 0.64]} \\
Qwen2.5-Math-7B-Instruct    & 0.199 \tiny{[0.14, 0.25]} & 0.063 \tiny{[0.05, 0.08]} & 0.155 \tiny{[0.12, 0.19]} & 0.289 \tiny{[0.23, 0.35]} \\
DeepSeek-Math-7B-RL         & 0.105 \tiny{[0.06, 0.15]} & 0.072 \tiny{[0.05, 0.09]} & 0.171 \tiny{[0.13, 0.21]} & 0.308 \tiny{[0.25, 0.37]} \\
Abel-7B-002                 & 0.045 \tiny{[0.02, 0.07]} & 0.052 \tiny{[0.04, 0.07]} & 0.124 \tiny{[0.09, 0.16]} & 0.224 \tiny{[0.17, 0.28]} \\
MAmmoTH2-7B-Plus            & 0.000 \tiny{[0.00, 0.00]} & 0.040 \tiny{[0.03, 0.05]} & 0.106 \tiny{[0.08, 0.14]} & 0.219 \tiny{[0.16, 0.28]} \\
\midrule
\textit{Qwen2.5-Math-1.5B} (smaller ref.) & 0.010 \tiny{[0.00, 0.02]} & 0.017 \tiny{[0.01, 0.03]} & 0.045 \tiny{[0.03, 0.07]} & 0.095 \tiny{[0.05, 0.13]} \\
\bottomrule
\end{tabular}
\caption{Open-weights baselines on the counterexample benchmark (n=201). All numbers Pass@$k$ with 95\,\% bootstrap CIs over problem ids. Sampled decoding uses $n=8$, temperature 0.7. Regenerated by \texttt{build\_camera\_ready\_tables.py}.}
\label{tab:oss-baselines}
\end{table*}

\begin{table*}[h]
\centering
\small
\begin{tabular}{lcccc}
\toprule
Model & Pass@1 (greedy) & Pass@1 (sampled) & Pass@3 (sampled) & Pass@8 (sampled) \\
\midrule
Gemini-3.1-Flash-Lite     & \textbf{0.577} \tiny{[0.51, 0.64]} & \textbf{0.565} \tiny{[0.50, 0.63]} & 0.638 \tiny{[0.57, 0.70]} & 0.687 \tiny{[0.62, 0.75]} \\
Gemini-3.5-Flash          & 0.493 \tiny{[0.42, 0.56]} & --- & --- & --- \\
Gemini-3-Flash-Preview    & 0.468 \tiny{[0.40, 0.54]} & 0.489 \tiny{[0.44, 0.53]} & \textbf{0.708} \tiny{[0.65, 0.76]} & \textbf{0.771} \tiny{[0.71, 0.83]} \\
Gemini-3.1-Pro            & 0.433 \tiny{[0.36, 0.50]} & --- & --- & --- \\
DeepSeek-Pro              & 0.428 \tiny{[0.36, 0.49]} & 0.437 \tiny{[0.38, 0.49]} & 0.551 \tiny{[0.49, 0.61]} & 0.622 \tiny{[0.55, 0.69]} \\
DeepSeek-Flash            & 0.423 \tiny{[0.35, 0.49]} & 0.432 \tiny{[0.38, 0.49]} & 0.562 \tiny{[0.50, 0.62]} & 0.637 \tiny{[0.57, 0.70]} \\
\bottomrule
\end{tabular}
\caption{Frontier commercial APIs on the counterexample benchmark (n=201). Dashes indicate sampled passes that were not run because token-budget constraints prevented an 8-sample run within the budget. Regenerated by \texttt{build\_camera\_ready\_tables.py}.}
\label{tab:api-baselines}
\end{table*}

\section{Witness-Schema Generator and Example Schemas}
\label{sec:appendix:schema}

The witness-schema stage (\S\ref{sec:method:schema}) is implemented in \texttt{tools/inject\_witness\_schema.py}. The tool iterates over verifier modules and queries a teacher model (DeepSeek via the OpenAI-compatible API client) with the verifier source code and the canonical witness expression, asking for a single-sentence Python-shape description of an acceptable witness, with a concrete example.

\paragraph{Representative schemas (one per subdomain).}
\begin{quote}\small
\textbf{GroupTheory.} ``Python tuple $(G, H)$ where both $G$ and $H$ are \texttt{sympy.combinatorics.PermutationGroup} objects constructed via SymPy group constructors (e.g., \texttt{SymmetricGroup(n)}, \texttt{AlternatingGroup(n)}, \texttt{DihedralGroup(n)}, \texttt{CyclicGroup(n)}). Example syntax: \texttt{(SymmetricGroup(3), DihedralGroup(4))}.''

\medskip

\textbf{RingTheory.} ``An \texttt{RModule} instance with attributes: \texttt{ring} (\texttt{CommutativeRing}) and \texttt{generators\_count} (\texttt{int}). Example: \texttt{RModule(Z, 2)}.''

\medskip

\textbf{FieldTheory.} ``A SymPy \texttt{Domain} object representing an integral domain that is not a field. Example: \texttt{ZZ} from SymPy.''

\medskip

\textbf{LinearAlgebra.} ``A SymPy \texttt{Matrix} object (square, non-real). Example: \texttt{Matrix([[0, 1], [-1, 0]])}.''

\medskip

\textbf{RealAnalysis.} ``Python tuple $(f, a, b, p)$ where $f$ is a SymPy expression (or string) in terms of \texttt{Symbol('x')}, $a$ and $b$ are real numbers $(a < b)$, $p$ is a real number.''

\medskip

\textbf{NumberTheory.} ``A Python int or \texttt{sympy.Integer} (single integer). Example: \texttt{5}.''
\end{quote}

The schemas are short (10--40 tokens) and add a small overhead to the user prompt (median schema length is well below the SFT user-prompt median of 120 tokens reported in the dataset stats).

\section{Artefact Licences}
\label{sec:appendix:licences}

All third-party artefacts are used for non-commercial research consistent with their release terms. Models: Qwen3-4B and Qwen2.5-Math (Apache~2.0), Gemma-3-4B (Gemma Terms of Use), DeepSeek-Math-7B-RL (DeepSeek Licence), Eurus-2-7B-PRIME, MAmmoTH2-7B-Plus and Abel-7B-002 (Apache~2.0). Libraries: vLLM, TRL, PEFT/LoRA, SymPy and Z3 (Apache~2.0 or MIT). Benchmarks: GSM8K (MIT), MATH-500 (MIT), MMLU (MIT), AIME-2024 (MIT). Commercial APIs were accessed under their providers' terms of service.

\textsc{SymCE} itself is released with the code under the MIT Licence and the corpus, verifier modules and human-audit annotations under CC~BY~4.0. The corpus consists exclusively of formal mathematical statements and their hypothesis-dropped mutations; it contains no human-derived text, no personal information, and no content from which personally identifying information could be derived.

\section{Eval-Suite Configuration and Reproducibility}
\label{sec:appendix:eval-config}

All evaluation in the main paper is orchestrated by the multi-model multi-benchmark suite runner
\pyrun{training.eval.run_suite}{\small\texttt{\ --config\ }}\pth{configs/eval/full_eval.yaml},
which iterates the cartesian product \{model\} $\times$ \{benchmark\} $\times$ \{prompt-style\} $\times$ \{greedy, sampled\} and writes one JSON per cell. Each per-cell JSON contains a \texttt{summary} block (Pass@$k$, by-domain, and (where applicable) subset metrics) and a \texttt{per\_record} block (one record per test problem with all sample-level fields: \texttt{completion}, \texttt{extracted}, \texttt{correct}, \texttt{reason}, \texttt{format\_ok}, \texttt{parse\_ok}, \texttt{is\_none}, oracle timing, etc.).

\paragraph{Reproducing the paper from the released artefacts.}
\begin{enumerate}\itemsep2pt\raggedright
\item \textbf{Every table.}\\
\pyrun{paper_preparation.scripts.build_camera_ready_tables}\\
regenerates all numbers in Tables~\ref{tab:main}, \ref{tab:trap}, \ref{tab:ood}, \ref{tab:multiseed}, \ref{tab:audit}, \ref{tab:scoring}, \ref{tab:perseed}, \ref{tab:memorisation}, \ref{tab:oss-baselines} and \ref{tab:api-baselines} directly from the raw per-record evaluation outputs, applying the answer-extraction rules of \S\ref{sec:experiments:benchmarks}. It is the single source of truth for this paper.
\item \textbf{Bootstrap CIs.}\\
\pyrun{paper_preparation.scripts.bootstrap_cis}\\
regenerates \pth{bootstrap_cis.csv} and \pth{bootstrap_deltas.csv}.
\item \textbf{Reason distribution (Fig.~\ref{fig:reason}).}\\
\pyrun{paper_preparation.scripts.eval_reason_dist}\\
regenerates the reason-counts table.
\item \textbf{All figures.}\\
\pyrun{paper_preparation.scripts.make_plots}{\small\texttt{\ --all}}\\
regenerates Figures~\ref{fig:reason}, \ref{fig:domain}, \ref{fig:training}, and \ref{fig:baselines} from the released CSVs.
\end{enumerate}

\paragraph{Dataset fingerprints.} Each suite run writes a \texttt{dataset\_fingerprint.json} file containing the SHA-256 hash and line count of every input JSONL used (the test set, the theorem-truth probe, and the external-benchmark sources). These fingerprints are sufficient to verify that all numbers in this paper were produced from the released splits.

\paragraph{Seeds.} Four training seeds (42, 19, 31 and 64) were used; the split seed is fixed at 42 throughout. The configs are sufficient to reproduce all training runs deterministically up to GPU non-determinism in the rollout backend (vLLM tensor-parallel sampling); the reported test-set numbers are robust to this residual non-determinism within the reported 95\,\% bootstrap CIs.

\end{document}